\documentclass[10pt]{article}
\usepackage[OE]{express}
\usepackage{ulem}

\newcommand{\rev}[1]{{\color{black} #1}}
\newcommand{\new}[1]{{\color{black} #1}}

\begin{document}
\title{Non-line-of-sight tracking of people at long range}

\author{Susan Chan,\authormark{1,**} Ryan E. Warburton,\authormark{1,**} Genevieve Gariepy,\authormark{1} Jonathan Leach,\authormark{1} and Daniele Faccio\authormark{1,*}}

\address{\authormark{1}Institute of Photonics and Quantum Sciences, Heriot-Watt University, David Brewster Building, Edinburgh, EH14 4AS, UK}

\email{\authormark{*}Corresponding author: d.faccio@hw.ac.uk}

\address{\authormark{**}These authors contributed equally.}

\begin{abstract}
A remote-sensing system that can determine the position of hidden objects has applications in many critical real-life scenarios, such as search and rescue missions and safe autonomous driving.  Previous work has shown the ability to range and image objects hidden from the direct line of sight, employing advanced optical imaging technologies aimed at small objects at short range.  In this work we demonstrate a long-range tracking system based on single laser illumination and single-pixel single-photon detection.  This enables us to track one or more people hidden from view at a stand-off distance of over 50~m.  These results pave the way towards next generation LiDAR systems that will reconstruct not only the direct-view scene but also the main elements hidden behind walls or corners.
\end{abstract}

\ocis{(150.5670) Machine vision, Range finding; (040.1345) Optoelectronics, Avalanche photodiodes (APDs).}


\section{Introduction}

Recent advances in light sensing and computational imaging technologies are providing solutions to the problem of looking around obstacles \cite{Kirmani:2009jr,Pandharkar:2011ha,Velten:2012ik,Gupta:2012dp,Heide:2013jc,Buttafava:2015ej,Laurenzis:2015dt,Gariepy:2015gi,Shrestha:2016id,Kadambi:2016fs,Klein:2016jk}.  In particular, devices that can detect the arrival of light at the single-photon level with extremely high temporal resolution have enabled the ranging and reconstruction of images of small, static hidden objects using methods based on laser-illuminated detection and ranging (LiDAR) \cite{Velten:2012ik,Buttafava:2015ej,Gariepy:2015gi}.  LiDAR is an active remote-sensing technique that uses the round-trip time-of-flight information of light signals backscattered from objects, typically in the line of sight, to determine their positions \cite{Wandinger:2005tl}.  LiDAR-based detection and ranging of objects hidden from view is achieved by illuminating the objects and detecting the backscattered signals via an intermediary scattering surface such as a wall or the floor.  These additional, intermediate scattering events and their isotropic nature greatly reduce the available signal for detection, leading to the need for long acquisition or processing times, and/or the need for advanced detection devices.

Buttafava \textit{et al.} recently studied the possibility of determining the full three-dimensional profile of static objects using a single-pixel single-photon avalanche diode (SPAD) and scanning laser illumination \cite{Buttafava:2015ej}.  Although the system did not have the necessary speed to track a moving object in real-time, \rev{single-pixel SPADs} do have a distinct advantage over SPAD cameras \cite{Niclass:2005cz,Zappa:2007fi,Richardson:2009ij,Charbon:2013jz} in that they provide close to $100\%$ coupling of light onto the sensitive detector area, compared to the few percent currently available in visible$-$near-infrared wavelength SPAD cameras \cite{Richardson:2009ks,Gariepy:2015ca}.

For some applications, object identification and position or motion information are the main points of interest; a three-dimensional reconstruction of the object is not crucial \cite{Pandharkar:2011ha,Gariepy:2015gi}.  Rather, the scenario may demand, first and foremost, knowledge of the presence and position of objects moving in a hidden environment.  Example applications include search and rescue, and autonomous driving.  Gariepy \textit{et al.} have shown that it is possible to detect and track a moving hidden object, albeit with no information of the object's form \cite{Gariepy:2015gi}.  Their setup uses a 1024-pixel SPAD camera \cite{Niclass:2005cz,Richardson:2009ij,Richardson:2009ks} to detect indirect light scattered back to the system from their target object around a corner.  Each detector pixel in the 32~$\times$~32~array has single-photon sensitivity and can time each photon's arrival with 110~ps resolution.  Tracking of the hidden object uses this precise photon-arrival timing information together with knowledge of the camera's field of view and the geometry of the setup in the line of sight.

\rev{Here we present a tracking system that extends beyond the lab-based near-range setups implemented thus far.  Our system enables the detection of people moving outside of the direct line of sight at stand-off distances greater than 50~m.}  The active imaging system uses one single-pixel SPAD detector and a single pulsed laser to ``look" down a corridor and around a blind corner.  In a first experiment, we demonstrate that we can accurately detect and locate a single person around the corner.  We then demonstrate that we can accurately do this for two hidden people.  The expected decrease in spatial resolution due to the use of a single- or few-pixel detection system compared to previous camera-based approaches is compensated by the arbitrarily large effective numerical aperture that is provided by using independently focusable detectors:  even moving objects at large distances can be located with high precision.  \rev{These results, therefore, show that single-photon counting technology can indeed perform outside the lab and over length scales that are relevant for large stand-off distance detection and location of hidden moving objects, such as humans, in real-life applications.}

\begin{figure}[t]
\centering
\includegraphics[width=0.85\columnwidth]{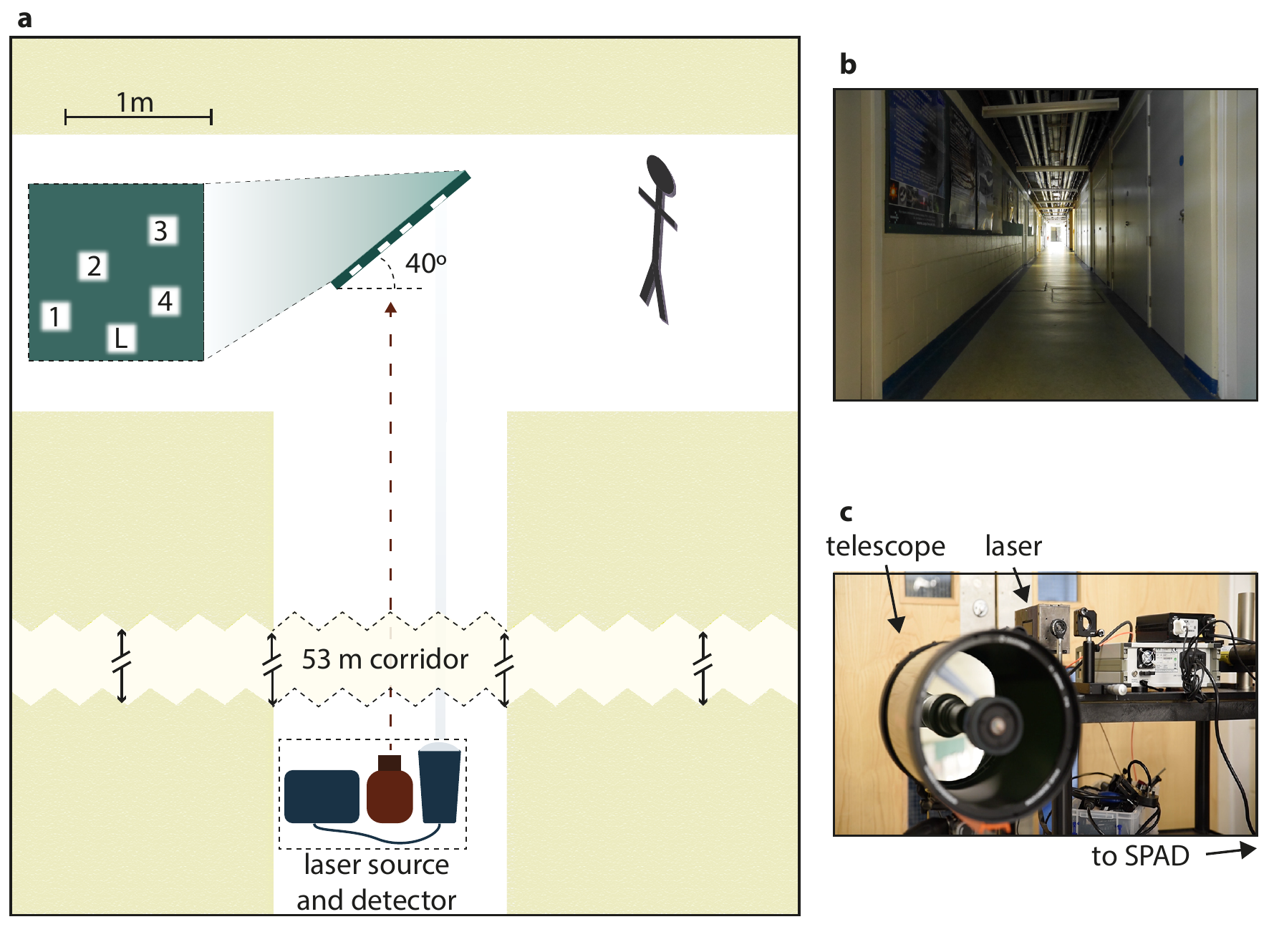}
\caption{Looking down the corridor.  We perform our experiment by looking down the corridor at a screen.  (a) shows a schematic overview of the geometry of the environment. The blind corners obstruct direct vision of the hidden person(s).  In a single measurement, we illuminate a single spot on the wall with a high-repetition pulsed laser and detect the light scattered back to the points on the wall which we image to our SPAD detector. ``L'' indicates the position of the laser on the screen and the numbered squares indicate positions imaged (one at a time) onto the single-pixel SPAD. (b) shows a photograph of the environment. Lighting conditions are as in the experiment. (c) shows a photograph of the equipment.}
\label{fig:setup}
\end{figure}

\section{Experimental setup}
In our experiments we use a corridor that forms a T-junction.  Figure~\ref{fig:setup} shows a schematic of the scenario as seen from above.  Compared to the experiment conducted by Gariepy \textit{et al.} where an intermediary scattering surface in the plane of the object's movement (the floor) was used, here we use a vertical surface (screen), as in previous work (see e.g. \cite{Kirmani:2009jr,Pandharkar:2011ha,Velten:2012ik,Gupta:2012dp}).  Our screen is a mobile chalkboard that we can freely position at the T-junction.  The laser and the field of view of the detector are directed onto viewing screens attached to the board to improve the signal level of our returns.

The transceiver comprises a pulsed laser diode (Picoquant LDH-P-FA-1530, 1532~nm peak wavelength, 40~MHz repetition frequency, $\sim400$~mW average power), \rev{a time-correlated single-photon counting \cite{Becker:2005wx} (TCSPC) module} (Picoquant PicoHarp 300), \rev{an InGaAs/InP SPAD} (ID Quantique ID230) and a commercial telescope (Celestron Schmidt-Cassegrain, 8-inch aperture, 80-inch focal length).  The returning light is coupled to the SPAD through a 62.5~$\mu$m diameter core fibre.  A 1500~nm long-pass filter was used.  \rev{The timing resolution of the detector is $\sim200$ ps.}

We send a train of light pulses from the transceiver onto our screen positioned at the far end of the corridor ($\sim53$~m away from the transceiver), as illustrated in Fig.~\ref{fig:setup}(a).  \rev{The 100~ps pulses from the portable laser} hit the wall with a \rev{$\sim7$ cm} beam diameter and scatter from the wall approximately as a spherical wavefront that propagates in all directions.  Some of this light reaches the hidden objects (persons) and is scattered back again towards the wall.  A discrete position with a spot size diameter of \rev{$\sim4$ cm} on the wall is imaged to \rev{the SPAD using the telescope.  Our TCSPC module} measures the photon arrival time (4~ps time binning) for the signal returning to the detector and a histogram is built up in one second of acquisition time over 40~million laser pulses.

\begin{figure}[b]
\centering
\includegraphics[width=0.85\columnwidth]{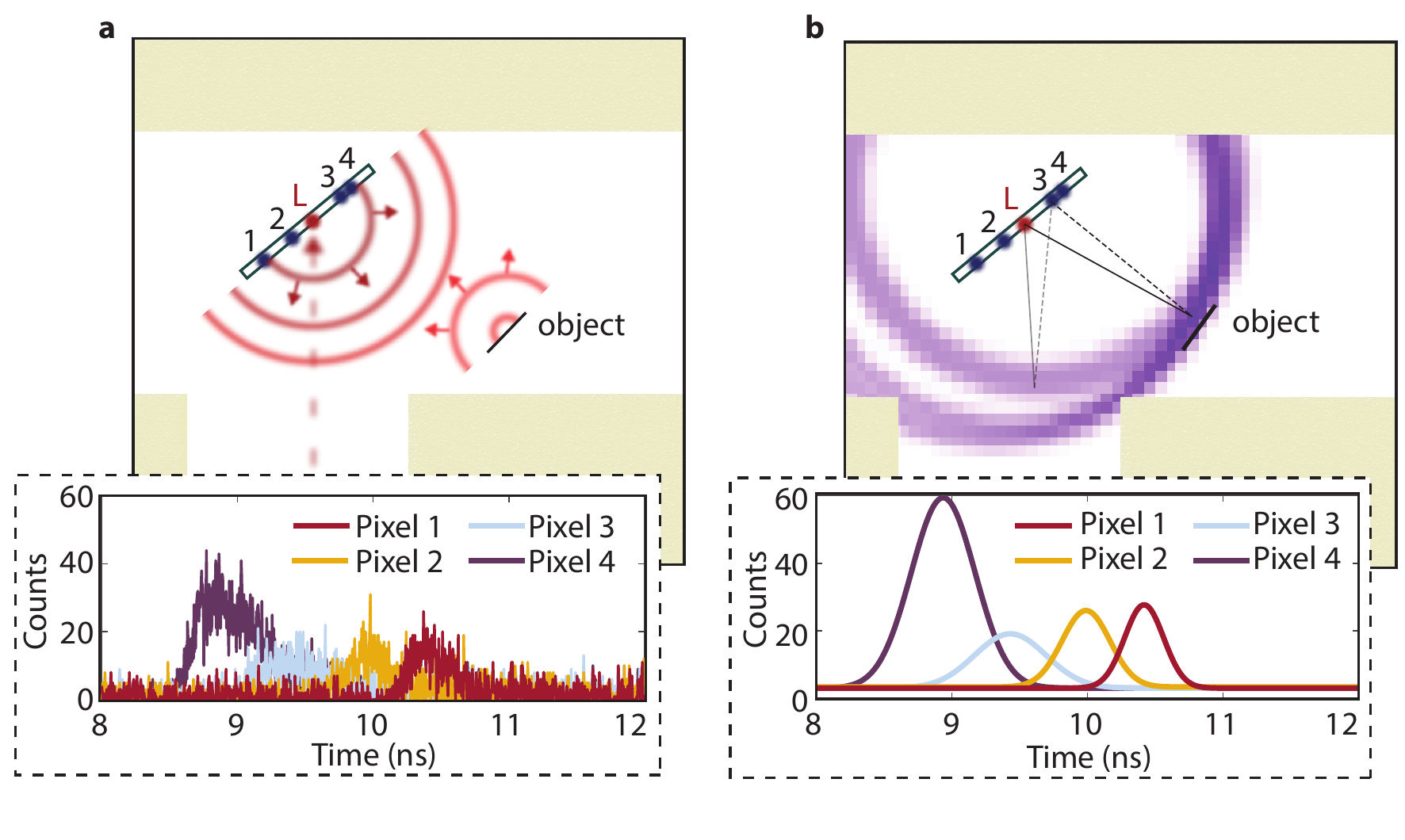}
\caption{Target-position retrieval.  (a)  Laser pulses incident on the wall scatter and propagate approximately as an isotropic spherical wavefront.  Light reaching an object scatters back in a similar fashion.  The inset histogram shows an example of the signal recorded by the single-pixel detector.  (b)  The time extracted from the peaks (of the Gaussian fits to the data, shown in the inset) provide the total distance travelled by the light, from the wall to the person and back to the wall, but not the actual path taken.  The bold and dashed lines in the schematic show two equivalent paths, corresponding to the time extracted from the histogram for one pixel position - there are infinite such paths that combined, form an ellipsoid of equally probable locations for the hidden object.  By repeating the measurement four times (looking at different positions on the screen as shown in Fig.~\ref{fig:setup}(a)) we obtain four ellipses that overlap in correspondence to the target position \new{in two dimensions}.}
\label{fig:hist}
\end{figure}

\section{Position retrieval}
In order to precisely locate a hidden object, we successively image four discrete positions on the screen to the SPAD as indicated in Fig.~\ref{fig:setup}(a).  This is equivalent to simultaneously imaging four detector positions to four SPADs with their associated collection optics; i.e. with four independent detectors and imaging systems, we would, in effect, be able to perform the total data acquisition in 1~s.  We use the temporal information between the laser signal and the SPAD detector signal for each of the four pixels (detector positions) to reconstruct the position of the hidden persons.  In our system, we can simply choose our point of reference (origin of the Cartesian coordinate system), for example the righthand-side corner of the T-junction, and we then use an extension of the approach presented in \cite{Gariepy:2015gi}.

To calibrate our measurements, we apply an offset to the histograms recorded for each pixel (detector position) and define the timing of all events relative to the start of the histograms.  We then limit the histogram to our time window of interest.  In our current method, we pre-acquire a signal of the background scene with the person(s) absent, for each detector pixel, and subtract this from the return signal for the corresponding pixel when we have the person(s) present in the scene.  Alternatively we can consider using the median of the histograms for each pixel \cite{Cutler:1998il,Cucchiara:2001is,Gariepy:2015gi} or frame-to-frame change detection \cite{Ralston:2010es,Pandharkar:2011ha} as a way to approximate the background signal.

For each detector pixel $i$, we perform a Gaussian fit on the histogram to locate the position of the peak(s) in the return signal corresponding to our scattering source(s) of interest (see insets in Fig.~\ref{fig:hist}).  Each fitted peak gives us our photon time-of-flight $t_i$, with uncertainty $\sigma_i$, that we measure between the initial instant when the train of laser pulses hits the wall at $r_l$ and the final instant when it has scattered to and from a hidden person at $r_o$, back into the field of view of the detector at $r_i$.  Using $t_i$, our method back-projects the photons onto the hidden scene.  We create an ensemble of discrete positions in Cartesian space for where our hidden person could be located.  Defining an appropriate scattering height $z$ for our person reduces the search space from a volume $(x,y,z)$ to an area $(x,y)$ with a significant increase in computation speed.  For each discrete position $r_o=(x_o,y_o,z)$, we then calculate a probability:

\begin{equation}\label{prob}
P_i(r_o) \propto \exp \left[ - \frac{(|r_o-r_l|+|r_o-r_i|-c t_i)^2}{2 \sigma_i^2} \right]
\end{equation}

The probability density that we obtain for our two-dimensional problem is maximised for a set of (x,y) that describes an ellipse with $r_l=(x_l,y_l,z_l)$ and $r_i=(x_i,y_i,z_i)$ as the foci, such that $|r_o-r_l|+|r_o-r_i|=c t_i$.  Here, $c$ is the speed of light.  Assuming independence between the measured times of flight $t_i$, the final distribution of $r_o$ is obtained by multiplying the probability density functions (PDFs) $P_i(r_o)$ associated with each pixel.  Since the counts in our target signal have contributions from the same scattering source, the resulting distribution will present a region of high probability where the individual densities overlap.

\section{Results}
\subsection{Tracking of a single person}
\begin{figure}[t]
\centering
\includegraphics[width=0.4\columnwidth]{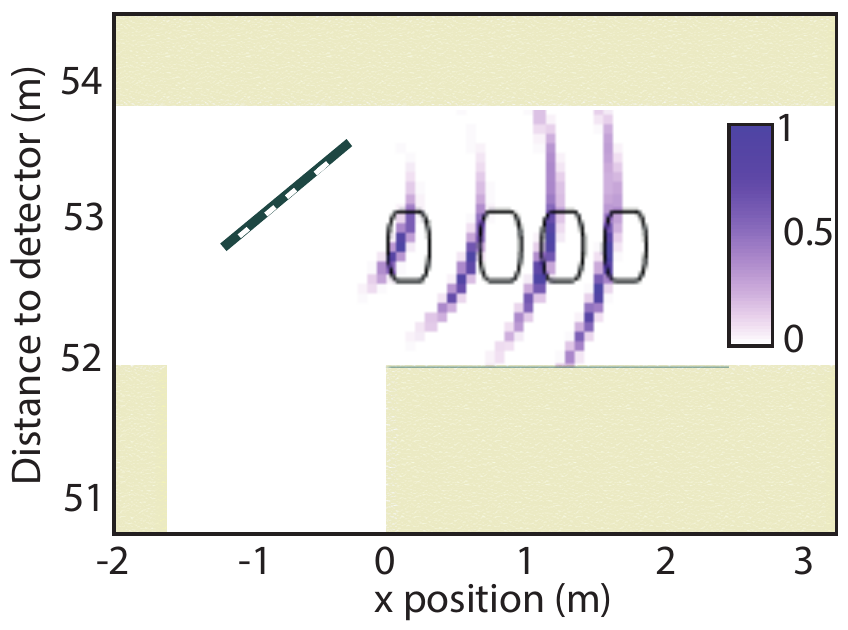}
\caption{Experimental results for non-line-of-sight tracking of a single person.  We perform tracking of a hidden person at distinct positions around a blind corner.  Each coloured curve in the graph is a joint probability distribution of the person's retrieved position, while the corresponding rectangle is their actual position during the measurement.}
\label{fig:single}
\end{figure}

A single person is standing around the righthand-side corner, hidden from the line of sight of our transceiver.  The screen is \rev{angled at $\sim40$ degrees to the corridor on the right to mimic, for example, an open door.}  The corridor lights are switched off yet full daylight illumination enters from lateral windows, as seen in Fig.~\ref{fig:setup}(b), however no significant differences were observed with the artificial corridor lights switched on or off. We acquire data for 1~s with the person located at each of four positions along the corridor, and we do this for all four pixels.  The signal of the background scene is also acquired for each pixel.  Figure~\ref{fig:single} shows the joint probability density functions (PDF) that we retrieve for the hidden person at each position.  These are overlaid with the ground truth.  The agreement between the PDFs and the person's true positions show that our system is able to locate a stationary person situated up to $\sim1.8$~m outside of the direct line of sight, and determine its position with an uncertainty that is always less than $\sim0.5$ m, although this increases the further away the person is from the screen.

\begin{figure}[b]
\centering
\includegraphics[width=.9\columnwidth]{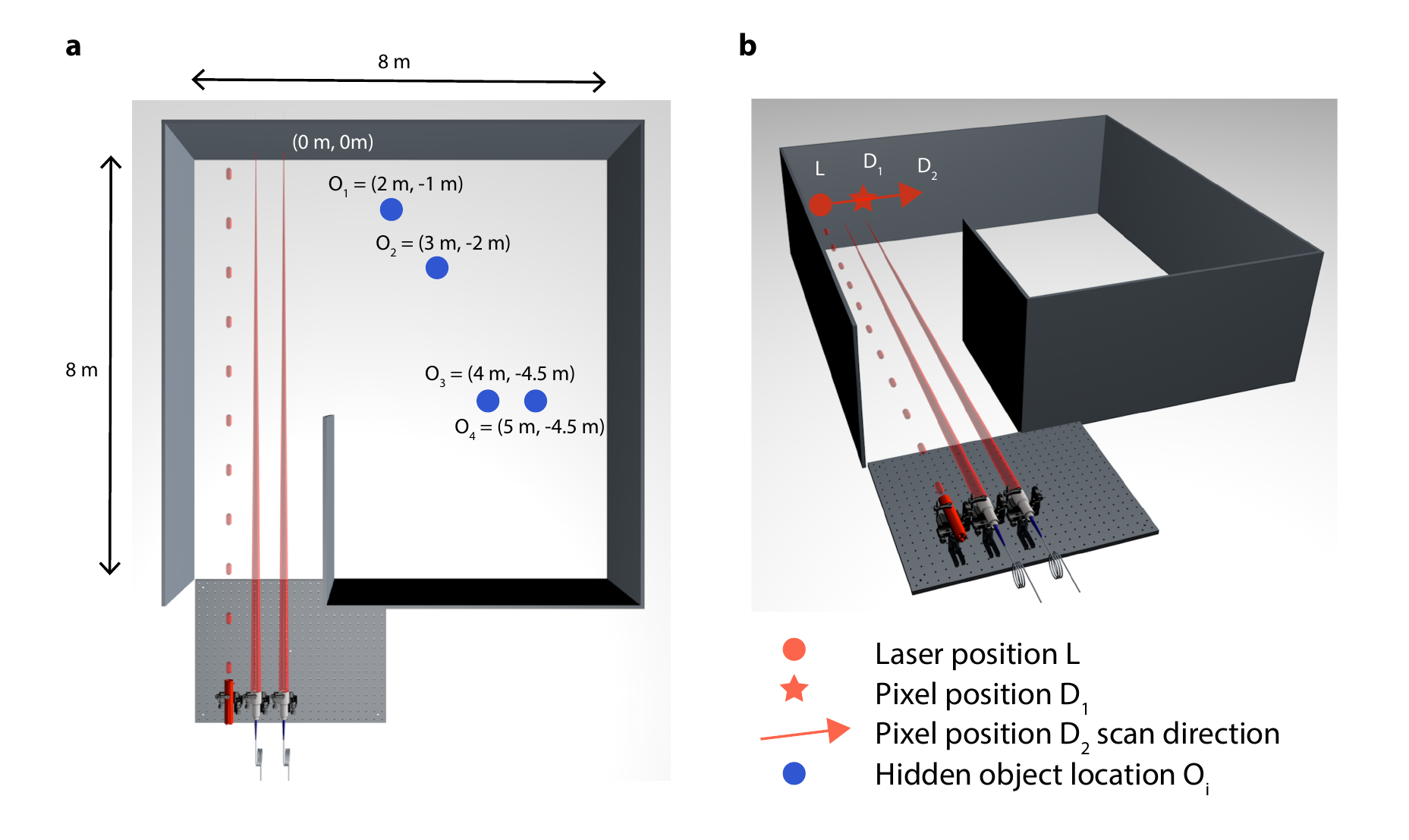}
\caption{\new{Simulation setup for a two-detector system.  To investigate the baseline distance between the detector positions on the screen, we simulate an experiment with a single laser and two detectors.  (a) and (b) show a schematic overview of the simulated environment in top-down and perspective view respectively.  Laser position "L" and pixel position "D$_1$" are fixed while pixel position "D$_2$" is scanned in the $x$-direction.  "O$_i$" indicates the location of the hidden object in each set of simulated measurements.}}
\label{fig:sim-setup}
\end{figure}

\begin{figure}[t]
\centering
\includegraphics[width=1\columnwidth]{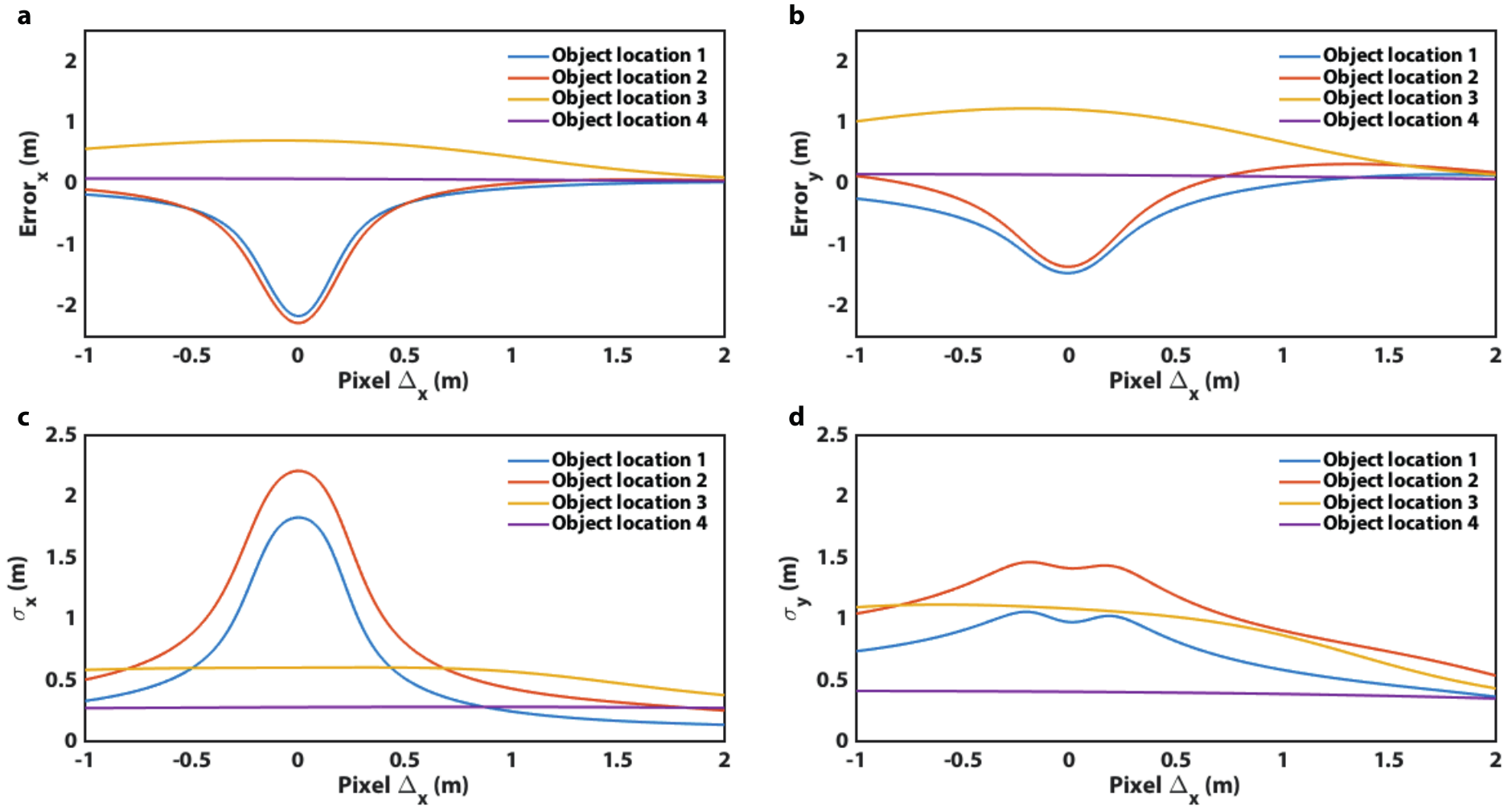}
\caption{\new{Simulation results for a two-detector system.  We investigate how separating out the pixels in one direction affects the accuracy and precision of the target position retrieval.  (a) and (b) show the respective change in accuracy of the retrieved $x$- and $y$-coordinates ("error$_x$" and "error$_y$") as the separation between pixels increases, while (c) and (d) show the change in precision in the $x$- and $y$-directions ("$\sigma_x$" and "$\sigma_y$") respectively.}}
\label{fig:sim-results}
\end{figure}

\new{In order to assess how location accuracy and precision varies with the baseline (i.e. the separation of pixel positions on the screen), we perform a computer simulation of the experiment for the simplified case where we have a single laser, two SPADs and a hidden object (see Fig.~\ref{fig:sim-setup}).  For four arbitrary object locations, we scan the pixel position of one of our detectors ($D_2$) in the $x$-direction to see how this affects the retrieved position.  The results indicate that a baseline of 1~m is sufficient to provide an average location accuracy of $\sim0.1$~m in the $x$-direction and $\sim0.25$~m in the $y$-direction (see Figs.~\ref{fig:sim-results}(a) and \ref{fig:sim-results}(b)), and an average location precision of $\sim0.4$~m and $\sim0.7$~m in the $x$- and $y$-directions respectively (see Figs.~\ref{fig:sim-results}(c) and \ref{fig:sim-results}(d)).  By introducing further pixels, a 1~m baseline will therefore be sufficient to guarantee a location precision comparable to the typical shoulder width of a human being.  In most cases, the precision increases very rapidly with increasing baseline but then tails off.  In other words, there is not much gain in going to large baselines - baselines of the order of 1~m are optimal in terms of requirements on the environment (access to a sufficiently large visible area) and location accuracy.  These results are verified in our experiments in which we have baselines of the order of 1~m.}

\subsection{Tracking of multiple persons}
\begin{figure}[h]
\centering
\includegraphics[width=0.85\columnwidth]{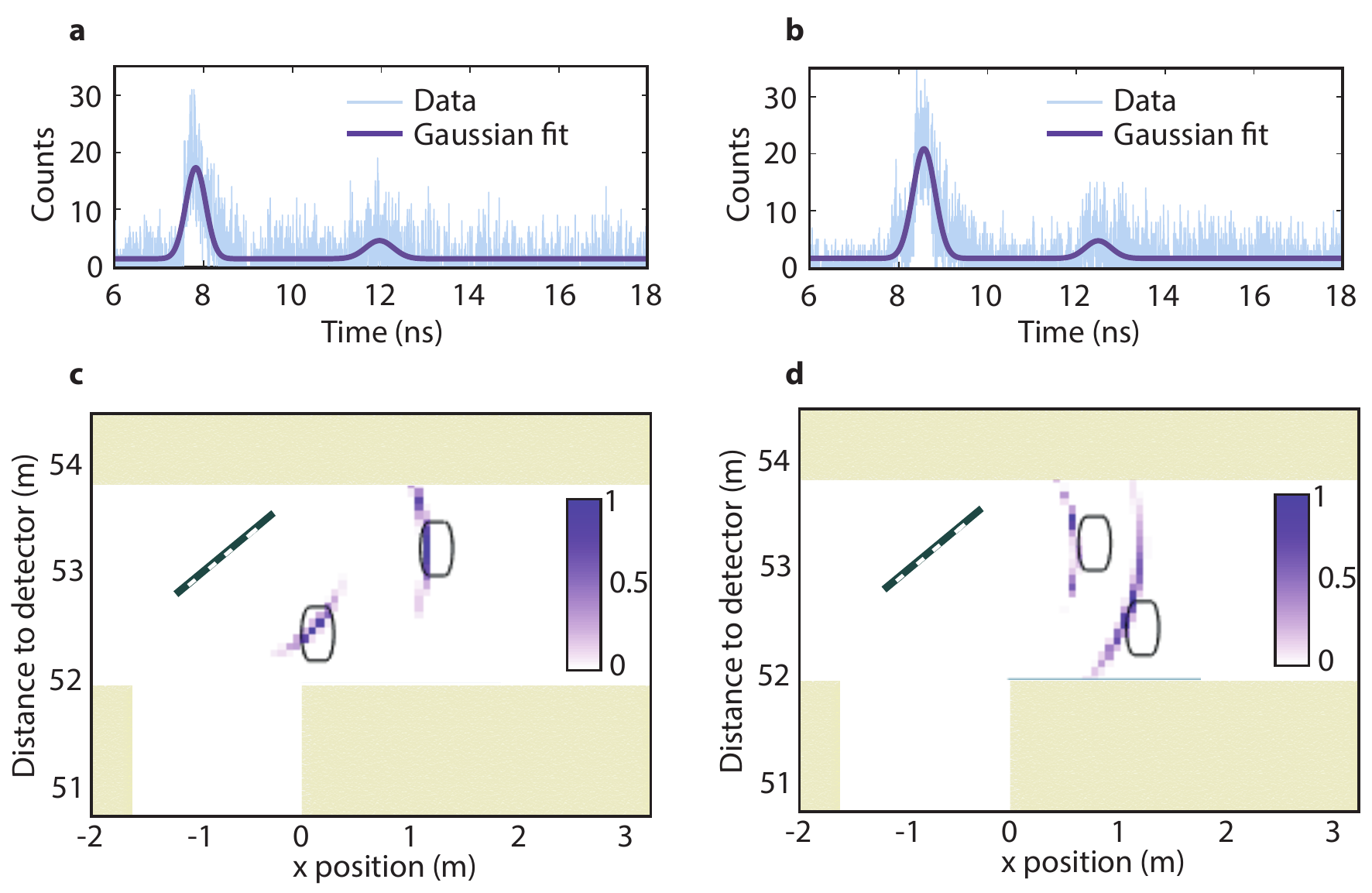}
\caption{Experimental results for tracking of multiple hidden persons.  We perform tracking of two people located around the same blind corner.  (a) and (b) show the raw data corresponding to just one of the four pixel positions used to identify and locate the two people, as shown in (c) and (d).} 
\label{fig:multiple-same}
\end{figure}

The previous experimental measurements are repeated with two people in the same hidden scene.   Again, data is acquired for 1~s for each detector position.  We show the results in Fig.~\ref{fig:multiple-same}.  Figures~\ref{fig:multiple-same}(a) and \ref{fig:multiple-same}(b) show examples of the actual data and corresponding Gaussian fits used for the retrieval, highlighting the presence and also clear difference between the return signals from each person. The retrieved positions, compared to the ground truth positions, are shown in Figs.~\ref{fig:multiple-same}(c) and \ref{fig:multiple-same}(d) respectively. As before, the positions of the people are always located to within $\sim0.5$~m for distances up to $\sim2$~m from the screen.

\begin{figure}[h]
\centering
\includegraphics[width=0.45\columnwidth]{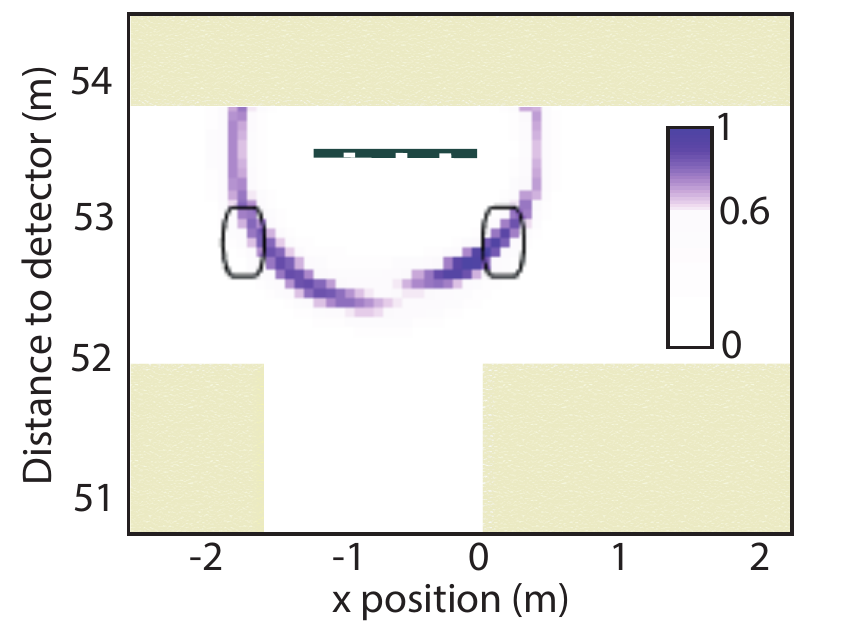}
\caption{Experimental results for detecting multiple hidden persons in less favourable conditions.  We perform detection of two hidden persons, each located around either corner of the T-junction. The coloured curves and corresponding rectangles show each person's retrieved position and actual position relative to the righthand-side corner.}
\label{fig:multiple-opp}
\end{figure}

We also perform a measurement with two people located on opposite sides of the T-junction.  We use only two detector positions and the screen is now placed at normal incidence with respect to the transceiver.  The retrieved positions are shown in Fig.~\ref{fig:multiple-opp}. The system is now retrieving the target positions with larger uncertainty and we observe that if either of the persons moved to further than $\sim2$~m away from the laser spot on the wall, the signal-to-noise ratio of the return signal degraded to the point that position reconstruction is no longer possible within the integration time.  \new{Preliminary measurements showed that person orientation has no notable effect on scattering.  Instead,} we attribute this largely to a non-uniform scattering distribution of the photons from the screen surface with a significantly lower distribution parallel to the wall itself (i.e. in the direction of the hidden persons in this configuration).  However, the presence of hidden people and their correct positions can still be identified by the system\new{, showing that no particular tuning of the angle of the screen is required.  That is,} the limitations observed here can, in principle, be overcome by using detectors with lower noise (the detector used here has a dark count rate of roughly 1000 counts/second and systems with a factor $\sim10\times$ less are available). The signal may also be increased by collecting the return scatter from a larger spot size, e.g. by optimising the collection optic numerical aperture. \rev{The spot size diameter could be increased by a factor $\sim2-3\times$ (therefore increasing signal by a factor $\sim10\times$) with respect to the current configuration without significantly compromising precision.  This would imply (when combined with lower noise detectors) a total increase in the signal-to-noise ratio of $\sim100\times$ and the potential to detect hidden objects that are located up to several meters behind the corner while maintaining the same 50~m stand-off distance.}  \new{The detected signal scales as $1/d^4$, where $d$ is the distance between the screen and the object(s) of interest.  With our current setup with the current technology, our method is limited to resolving two simultaneous objects in the hidden scene; any increase will require increased temporal resolution.}

\section{Discussion}
Photons returning via objects located outside of the line of sight undergo \rev{three or more scattering events}, yet the timing information of return signals can be used to accurately retrieve the positions of these objects.    The flexibility of single-pixel detection provides an increased field of view that allows detection and tracking with precision at long range, $>50$ m.  The use of single-pixel detectors also has the advantage of high detection efficiency.  The results from our measurements show that by using a few single-pixel SPADs in parallel and TCSPC, real-time tracking at large stand-off distances is possible.  The ability to perform non-line-of-sight detection at long range, using experimental components that can be made both compact and portable, takes us one step closer to developing a solution that is usable for real-life scenarios. An interesting possibility also is to use just one single pixel: while this will not provide sufficient information to locate the exact position of a hidden object, it is sufficient to identify the existence of a moving object, its distance from the laser beam spot, its direction of motion and velocity.

\section*{Funding}
This work was funded by the European Research Council under the European Union's Seventh Framework Programme (FP/2007-2013)/ERC GA 306559, by the Engineering and Physical Sciences Research Council (EPSRC, UK; grants  EP/M006514/1, EP/M01326X/1), and by the Defence Science and Technology Laboratory's Centre for Defence Enterprise.

\section*{Acknowledgments}
We would like to thank Alessandro Boccolini for the photographs used in this paper.   The authors acknowledge discussions with Ken McEwan.  All experimental data is available at http://dx.doi.org/10.17861/8a0669d3-8f9b-4d8a-b204-f980f5d57c29.

\end{document}